\documentclass[letterpaper]{article} 
\usepackage{aaai2026}  
\usepackage{times}  
\usepackage{helvet}  
\usepackage{courier}  
\usepackage[hyphens]{url}  
\usepackage{graphicx} 
\usepackage{natbib}  
\usepackage{caption} 
\usepackage{algorithm}
\usepackage{algorithmic}

\usepackage{newfloat}
\usepackage{listings}
\DeclareCaptionStyle{ruled}{labelfont=normalfont,labelsep=colon,strut=off} 
\floatstyle{ruled}
\newfloat{listing}{tb}{lst}{}
\floatname{listing}{Listing}
\title{VitalDiagnosis: AI-Driven Ecosystem for 24/7 Vital Monitoring and Chronic Disease Management}
\author{
    Zhikai Xue\textsuperscript{\rm 1}, Tianqianjin Lin\textsuperscript{\rm 2}, Pengwei Yan\textsuperscript{\rm 2}, Ruichun Wang\textsuperscript{\rm 1}\\
    Yuxin Liu\textsuperscript{\rm 3}, Zhuoren Jiang\textsuperscript{\rm 2}\thanks{Corresponding authors.}, Xiaozhong Liu\textsuperscript{\rm 1*}\\
}
\affiliations{
    \textsuperscript{\rm 1}Worcester Polytechnic Institute\\
    \textsuperscript{\rm 2}Zhejiang University\\
    \textsuperscript{\rm 3}University of Delaware\\
    \{zxue1, rwang12, xliu14\}@wpi.edu, \{lintqj, yanpw, jiangzhuoren\}@zju.edu.cn, cindyliu@udel.edu
}

\usepackage{bibentry}

\begin{document}

\maketitle

\begin{abstract}
Chronic diseases have become the leading cause of death worldwide, a challenge intensified by strained medical resources and an aging population. Individually, patients often struggle to interpret early signs of deterioration or maintain adherence to care plans. In this paper, we introduce VitalDiagnosis, an LLM-driven ecosystem designed to shift chronic disease management from passive monitoring to proactive, interactive engagement. By integrating continuous data from wearable devices with the reasoning capabilities of LLMs, the system addresses both acute health anomalies and routine adherence. It analyzes triggers through context-aware inquiries, produces provisional insights within a collaborative patient–clinician workflow, and offers personalized guidance. This approach aims to promote a more proactive and cooperative care paradigm, with the potential to enhance patient self-management and reduce avoidable clinical workload.

\end{abstract}

\begin{links}
    \link{Demo Repository}{https://tinyurl.com/5n83hcrz}
\end{links}

\section{Introduction}
\noindent Chronic diseases, including cardiovascular conditions, diabetes, and stroke, have become the leading causes of death worldwide \cite{stuckler2008population,strong2005preventing}. Unlike common illnesses, managing chronic diseases is a long-term, complex process that requires sustained medical engagement and considerable resource allocation, contributing to growing healthcare demands \cite{wagner1998chronic,wagner1996organizing,tan2024rpm}. This pressure, intensified by an aging population, has strained medical systems and led to delays in timely care delivery \cite{smith2017disparities}, limiting opportunities for early intervention that could mitigate severe complications \cite{adigun2019urgent}. At the individual level, patients often lack the clinical literacy needed to self-manage their conditions or recognize early signs of deterioration \cite{lu2025digitaladherence}. 
As a result, healthcare systems remain largely reactive, struggling to keep pace with the escalating burden of chronic illness \cite{choi2024living,gbd2025chronic}.

To address these challenges, we introduce \textbf{VitalDiagnosis}, an ecosystem of specialized LLMs designed to shift chronic care from passive monitoring to proactive engagement. 
The entire workflow is powered by Unified Memory Core, an architecture in which a central Memory MiniLLM maintains deep, continuous context by integrating medical knowledge bases and patient-related assets with an adaptable parametric memory composed of shared and personalized LoRAs. 
The process begins with a lightweight multimodal Monitoring MiniLLM that interprets wearable data to produce clinically relevant narratives and identify potential trigger events. 
The system then activates its dual-track framework through a Domain LLM specialized in clinical inquiry: for potential outliers, it initiates a dialogue to contextualize the event; for routine management, it engages patients to monitor adherence. 
The resulting insights are synthesized into provisional clinical assessments and actionable recommendations, which are circulated within a collaborative patient–clinician workflow for timely review. 
This proactive, data-driven approach aims to promote more anticipatory chronic care, with the potential to enhance self-management and reduce avoidable clinical workload.

\section{Related Work}
\noindent While LLMs show significant potential in healthcare \cite{goyal2024healaidoc,li2025llm-chronic,zhou-etal-2025-mam}, their integration with wearable devices remains nascent \cite{fang2024physiollm,ferrara2024llmwearable,zhang2025sensorlm}. Prior work in wearable-based chronic care has largely been confined to passive, threshold-based systems that trigger simple alerts for outliers \cite{de2024mobile,xie2021integration,subramanian2020precision}. These approaches fundamentally lack the capacity for nuanced, interactive investigation of anomalies and fail to provide continuous, adaptive support for personalized routine adherence \cite{zhou-etal-2025-improving}. In contrast, our VitalDiagnosis ecosystem addresses these gaps by introducing a dual-track framework that leverages the synergy of LLMs and wearables to support both real-time, interactive triage and proactive routine care within a collaborative patient–clinician loop.

\section{VitalDiagnosis Ecosystem}

\begin{figure*}[t]
  \centering
  \includegraphics[width=\textwidth]{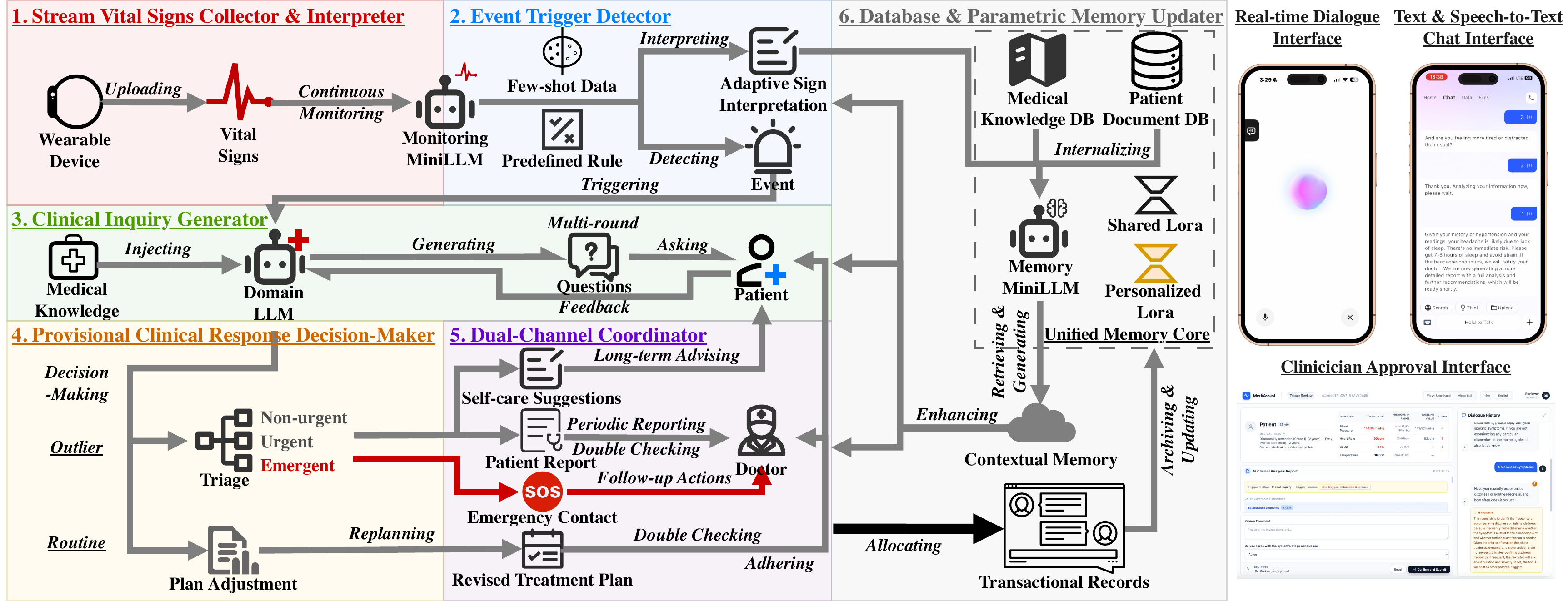}
  \caption{The framework and interfaces of the VitalDiagnosis ecosystem.}
  \label{fig:model}
\end{figure*}

\noindent VitalDiagnosis integrates LLMs with wearables into a dual-track framework for chronic disease management. By supporting both interactive triage of anomalies and proactive adherence monitoring, it establishes a collaborative patient–clinician loop that promotes more anticipatory care. Powering this workflow is the Unified Memory Core. 
Mediated by a central Memory MiniLLM, it combines persistent medical-knowledge databases and patient assets with a parametric memory of shared and personalized LoRAs \cite{zhang2024param-memory}, producing episodic or structured summaries that condition all downstream processes.
\footnote{We utilize LLMs of varying sizes: Memory MiniLLM (4B), Monitoring MiniLLM (1.7B), and Domain LLM (14B).}

\subsubsection{Stream Vital Signs Collector \& Interpreter}
The workflow begins by continuously collecting vital sign streams from wearable devices. A lightweight multimodal Monitoring MiniLLM then interprets these raw, variable-length signal segments into concise, clinician-readable narratives, supplying the initial context for all subsequent tasks.

\subsubsection{Event Trigger Detector}
Next, the system identifies clinically relevant triggers by inferring the patient's state using track-specific logic. It applies rule-based thresholds and model-based inference for anomalies, while also scheduling periodic checks for routine care. The system then routes each event to either \emph{outlier-detection} or \emph{routine-adherence} logic, producing a risk-graded trigger to guide the next stage.

\subsubsection{Clinical Inquiry Generator}
Upon event triggering, a scene-specialized Domain LLM, adapted with LoRA on clinician-annotated simulated and rewritten cases, initiates a brief Q\&A session with the patient, conditioned on contextual memory. For outlier events, it explores symptoms and contributing factors; for routine management, it assesses adherence and notes potential barriers. The inquiry remains minimal, concluding once adequate information is obtained.

\subsubsection{Provisional Clinical Response Decision-Maker}
Using interpreted events, inquiry results, and contextual memory, the system produces a \emph{provisional} clinical response. For outliers, it provides severity-aware triage with potential factors and next steps; for routine care, it summarizes adherence and suggests refinements. All recommendations follow a tiered approval process, with high-risk items requiring explicit review and low-risk items eligible for deferred confirmation.

\subsubsection{Dual-Channel Coordinator}
The system then coordinates all communication, translating provisional judgments into audience-specific reports bounded by visibility constraints. It delivers triage-aware guidance to patients while presenting providers with a concise event summary and any flagged high-risk proposals for review. For routine management, it offers patients adherence suggestions and supplies providers with periodic digests for confirmation.

\subsubsection{Database \& Parametric Memory Updater}
To complete the loop, the system updates its memory with all interactions and outcomes. The memory database maintains a short-term rolling snapshot of recent activity, while confirmed facts are promoted to long-term storage with full traceability. The parametric memory (LoRA modules) is updated more selectively on stable patterns emerging from the long-term data, ensuring the system is continuously refined through accumulated interactions and expert oversight.

\section{Conclusion}
\noindent In conclusion, VitalDiagnosis presents an ecosystem that integrates LLMs with wearables to support a dual-track framework for chronic disease management. By enabling both interactive triage of health anomalies and proactive monitoring of routine adherence, it fosters a collaborative patient–clinician loop that promotes more anticipatory care. This scalable, proactive paradigm has the potential to enhance patient self-management, ease provider workload, and broaden access to continuous, personalized support. 
To further assess its real-world performance and contribute to the community, VitalDiagnosis is currently undergoing pilot studies with medical institutions, and a curated, clinician-annotated dataset will be made publicly available.

\bibliography{aaai2026}

\end{document}